\documentclass[11pt]{article}
\usepackage[final]{acl}

\usepackage{times}
\usepackage{latexsym}
\usepackage[T1]{fontenc}
\usepackage[utf8]{inputenc}
\usepackage{microtype}
\usepackage{graphicx}
\usepackage{booktabs}
\usepackage{multirow}
\usepackage{array}
\usepackage{amsmath}
\usepackage{amssymb}
\usepackage{xcolor}
\usepackage{makecell}
\usepackage{pgfplots}
\pgfplotsset{compat=1.18}

\newcommand{\best}[1]{\textbf{#1}}        
\newcommand{\bestopen}[1]{\underline{#1}} 

\title{Cloud and On-Premises Deployment of Uzbek Legal RAG via Targeted Retriever Fine-Tuning}

\author{Tatul Danielyan \quad Mariam Avetisyan \quad Hrant Davtyan \\
  Metric AI Lab \\
  \texttt{\{tatul, mariam, hrant\}@metricailab.com}}

\begin{document}
\maketitle

\begin{abstract}
Deploying large language models for legal question answering raises challenges
that general-purpose leaderboards do not capture, particularly for low-resource languages and under hard operational constraints. We report on building and operating a retrieval-augmented (RAG) legal assistant for Uzbek that must run in two regimes: a managed cloud service that maximizes answer quality within a per-token cost ceiling, and an on-premises deployment for clients whose legal data may not leave their infrastructure, restricting us to open-weight models on limited local hardware under latency constraints. Because no evaluation existed for this setting, we build two domain benchmarks: a retrieval benchmark of 178 expert-annotated legal queries with gold provision spans, and an end-to-end benchmark of 504 expert-curated question--answer pairs scored by an LLM judge whose ratings we validate against human judgments and against an independent-family judge. Applying these benchmarks under each regime, we find the open-versus-proprietary gap is small and cheaply closed by fine-tuning. Therefore, we train UTE-1, which is a state-of-the-art text embedder among open models for Uzbek. We also demonstrate that closing the performance gap via fine-tuning is both impractical due to the intensive hardware demands of long-context legal Q\&A and unnecessary, given that legal acts change frequently. We support this by reporting a negative result from a QLoRA experiment. We distill practical guidance for similar deployments, drawn from a system serving real users in production. We release our benchmarks, evaluation code and the fine-tuned embedder (UTE-1) \href{https://metric-ai-lab.github.io/Uzbek-Legal-RAG/}{at this https URL} to support future work on low-resource legal NLP.
\end{abstract}

\section{Introduction}

Large language models (LLMs) are increasingly deployed in legal and governmental
settings, where citizens and professionals ask natural-language questions and
expect answers grounded in the applicable statutes. Such systems are typically
built as retrieval-augmented generation (RAG) pipelines
\citep{lewis2020rag,gao2023ragsurvey}: a retriever selects relevant legal text,
and an LLM composes an answer conditioned on it. RAG is a natural fit for the
legal domain because the law is large, frequently amended, and demands
attribution to specific provisions.

Most published guidance on building such systems is implicitly written for
high-resource languages and a single, unconstrained deployment target. Real
deployments are neither. We describe a production legal assistant for
\textbf{Uzbek}, a Turkic language spoken by tens of millions but under-served by
NLP resources, grounded in the national legal corpus (\texttt{lex.uz}). The system
is commercially deployed and serves real users; per the double-blind policy we
omit the product and organization names.

Two facts make model selection for this system non-obvious. First, the same
assistant ships in \emph{two deployment regimes} with conflicting constraints. The \emph{cloud} regime is a multi-tenant SaaS
offering: any model is permissible, but a per-token cost ceiling rules out the
most expensive frontier APIs, and the objective is to maximize quality within
that budget. The \emph{on-premises} regime serves enterprise and public-sector
clients whose legal documents \emph{cannot leave their infrastructure}; data
sovereignty restricts us to open-weight models, which must then run on the
client's limited local hardware under latency constraints. Second, general
multilingual leaderboards poorly predict quality here: strong aggregate ability
does not imply strong Uzbek \emph{legal} retrieval or answer quality, and the
public Uzbek leaderboards that exist measure generic language ability, not
grounded legal QA. Principled model choice therefore requires task- and
language-specific evaluation that did not previously exist.

We make three contributions.

\noindent\textbf{(1) A deployment case study with a transferable decision
procedure.} We cast cloud and on-premises model selection as two constrained
optimization problems and show how one evaluation suite drives both, yielding two
deployed model stacks.

\noindent\textbf{(2) Evaluation infrastructure for Uzbek legal QA.} We release a
178-query retrieval benchmark with gold provision spans and a 504-item end-to-end
QA benchmark, plus a \emph{human-validated} LLM-as-judge protocol cross-checked
against an independent-family judge---the linchpin of any such study. Both benchmarks and evaluation code are released.

\noindent\textbf{(3) The finding that, under deployment constraints, the
retriever is where to invest.} Existing legal-RAG work fine-tunes the generator \cite{ma2025sdd, octadion2024asvri}, and parallel work \cite{butler2026legal} finds the retriever dominant but does not fine-tune it. We fine-tune an open embedder to a state-of-the-art
open result for Uzbek (cheaply closing roughly half the in-domain gap to proprietary embeddings),
and report a \emph{negative} result: fine-tuning the generator under the same
constraints is both costly to pursue and, given that laws change and multilingual
LLMs keep improving, unnecessary. The heuristic---\emph{fine-tune the retriever,
rent or swap the generator}---is the paper's central, reusable lesson. The text embedder is released publicly to further facilitate NLP applications in Uzbek language.

\section{System and Deployment Setting}
\label{sec:system}

\paragraph{Pipeline.} The assistant is a RAG question-answering
pipeline. Offline, we segment the national legal corpus into provision-level
chunks, prepend each with its hierarchical headings (code/chapter/article) so
locally ambiguous text stays interpretable, embed them, and index them in a vector
database (Weaviate). At query time we normalize (and where needed translate) the
question, then retrieve with \emph{hybrid} search---dense cosine similarity
\citep{karpukhin2020dpr} fused with lexical matching---since legal queries hinge
on exact terms (article numbers, statute names) that pure semantic search misses.
We prompt an LLM with the retrieved provisions and a legal-assistant instruction
to generate an answer citing the provisions it used. The production path retrieves
$k{=}50$ chunks, so the retriever must ensure the governing provision is present in
a 50-passage window (engineering details in Appendix~\ref{app:repro}).

\paragraph{Why two models, and why they matter unequally.} A RAG pipeline has
two learnable components, the \emph{embedder} (retriever) and the
\emph{generator}. Their failure modes are asymmetric: if the embedder fails to
surface the governing provision, no generator can recover---the answer is
confidently wrong or unsupported---whereas a capable generator mostly needs to
read, synthesize, and attribute text it is given. This asymmetry motivates
treating the embedder as a first-class object of optimization, a theme we return
to in Sections~\ref{sec:onprem} and~\ref{sec:discussion}.

\paragraph{Constraints.} The retrieved context is large: in production, the $k{=}50$
provisions for a query span roughly \textbf{55k--90k tokens}. Two domain factors
drive this: the same statute is cited across many articles, so a small top-$k$
returns cross-references rather than governing text and high $k$ is needed to
capture them all; and statutory tables too large to embed as one unit are indexed
as several chunks but re-expanded in full when any is retrieved. This long-context
regime shapes both cost and latency. In the cloud regime, per-query
cost is dominated by the generator's input tokens, so the cost ceiling
effectively excludes frontier ``pro''-tier APIs and admits only efficient
``flash''-tier or open models. In the on-premises regime, the sovereignty
constraint removes all API models, and latency is driven both by prefill over the
large context and by decoding long legal answers that may require multi-step
reasoning over several provisions---so prompt processing \emph{and} decode time
shape the user-visible wait.

\section{Benchmarks and Evaluation Methodology}
\label{sec:eval}

Because no public benchmark targets Uzbek legal RAG, and because the central
risk in this study is trusting an unreliable evaluator, we invested first in
evaluation. We build two complementary benchmarks and validate the automated
judge against humans.

\subsection{Retrieval benchmark}
\label{sec:retrbench}
We construct a retrieval benchmark of 178 human-written and human-annotated legal questions\footnote{All questions are in Uzbek with Latin script.}. For each question, domain experts identified the exact provision text required to answer it and recorded those passages as gold targets. We perform in-batch evaluation and report \emph{top-5 retrieval accuracy}: whether at least one gold provision appears among the five highest-ranked chunks. To situate our embedders externally, we also evaluate on the Uzbek subset of MTEB \citep{muennighoff2023mteb} (127 queries over 1{,}263 candidates, nDCG@5). The two sets are independent in content and scoring and together cover 305 queries, measuring in-domain legal and general Uzbek retrieval respectively.

\subsection{End-to-end QA benchmark}
\label{sec:qabench}
We curate 504 legal question--answer pairs spanning 22 legal domains \footnote{312 questions are in Uzbek with Latin script, 91 are in Cyrillic Uzbek, and 101 are in Russian}. 384 out of 504 (76\%) are composed by legal experts (both questions and answers), 60 are sampled at random from search-style production queries (answers are expert-annotated), 60 are sampled at random from upvoted interactions (answers are LLM-drafted then reviewed, reformulated, and approved by experts).

Each item is scored by running the \emph{entire production pipeline}: the question is retrieved against the live vector index ($k{=}50$) and answered by the generator, and the generated answer is compared against the reference. We score four quality dimensions with an independent LLM judge: \emph{completeness}, \emph{contextual accuracy} (faithfulness to retrieved provisions), \emph{hallucination-free}, and \emph{legal comprehension}, and report an \emph{overall} score. Because it evaluates the deployed pipeline rather than models in isolation, the benchmark directly measures the quantity we care about and lets us vary the embedder and generator  independently (Section~\ref{sec:onprem}).

\subsection{LLM judge}
\label{sec:judge}
Our headline QA metric is produced by an LLM judge. Using judge from the same family as responder LLM creates a risk for self-preference bias: LLM evaluators tend to favor their own family \citep{panickssery2024self,zheng2023judge} potentially inflating the shared-family generator's score. We therefore pick a capable LLM as judge which is from a completely different family than evaluated LLMs (\texttt{Muse Spark 1.1}).

We compare the chosen independent judge against human ratings. On a set of 52 questions scored on a 1--10 scale by domain experts, our judge attains Spearman $\rho{=}0.7032$ and a mean absolute error of 1.29 points against the expert scores.

\section{Cloud Deployment: Quality Within a Cost Ceiling}
\label{sec:cloud}

In the cloud regime any model is admissible in principle, so selection reduces to
maximizing quality subject to the cost ceiling.

\paragraph{Embedder.} Table~\ref{tab:embed} reports retrieval quality for
fourteen embedders. The proprietary \texttt{gemini-embedding-001} is best on our
in-domain benchmark (0.961 top-5) and on MTEB (0.906 nDCG@5), with comfortable
margins. As cloud embedding costs are a small fraction of total cost, we deploy
it for the cloud service.

\paragraph{Generator.} We pre-filter LLM candidates by three criteria: (i) sufficient general Uzbek ability, using a public Uzbek leaderboard (UzLib; full table in Appendix~\ref{app:uzlib}); (ii) sufficient general capability, using public aggregate signals (e.g., arena rankings and MMLU-Pro); and (iii) price below our per-token ceiling.\footnote{Our ceiling is \$3 per million input tokens and \$15 per million output tokens---a business decision balancing answer quality against competitive end-user pricing and local purchasing power. Because retrieved legal context runs to 55--90k tokens per query (\S\ref{sec:system}), per-query cost is dominated by input tokens, so the input-token price is the binding figure; the answer itself is a small fraction.} This excludes the most expensive frontier APIs (e.g., the Claude, Gemini-Pro, and GPT-Pro tiers) despite their strong general scores. Among the in-budget candidates we then rank by end-to-end quality on our QA benchmark (Table~\ref{tab:qa}). \texttt{gemini-3-flash} is the strongest in-budget generator (0.850 overall); the more expensive reference models we evaluate for context (e.g., \texttt{claude-sonnet-4-6}, \texttt{gpt-5.4}) do not surpass it on this task. We therefore deploy \texttt{gemini-embedding-001} $+$ \texttt{gemini-3-flash} for the cloud service.

\paragraph{Cost and quality together.} The cost ceiling is not a quality sacrifice. Plotting per-model inference cost against QA quality (Appendix~\ref{app:cost}, Figure~\ref{fig:pareto}), \texttt{gemini-3-flash} sits at the quality-maximizing vertex of the cost--quality Pareto frontier and \emph{strictly dominates} every frontier API: it is $2.1\times$, $4.0\times$, and $6.6\times$ cheaper than \texttt{grok-4.3}, \texttt{gpt-5.4}, and \texttt{claude-sonnet-4-6} respectively, while scoring higher than all three. For this task, ``rent the flash tier, not the frontier'' is the cost--quality-optimal choice, not a budget compromise.

\paragraph{Production corroboration.} The deployed cloud stack serves
$\approx$30k questions per month. Over 479 in-product votes, 78\% were positive
(375/479), corroborating the offline QA ranking that selected this stack. The
on-premises deployments surface \emph{no} such telemetry by design: the same data
sovereignty that motivates them withholds the feedback signal the cloud service
relies on, a tradeoff we revisit in \S\ref{sec:discussion}.

\begin{table*}[t]
\centering
\small
\setlength{\tabcolsep}{6pt}
\begin{tabular}{@{}llcccc@{}}
\toprule
& \textbf{Embedder} & \textbf{Open} & \textbf{Retr.\ Top-5} & \textbf{MTEB nDCG@5} & \textbf{Avg} \\
\midrule
1  & \texttt{gemini-embedding-001} (3072d) & --- & \best{0.961} & \best{0.906} & \best{0.934} \\
2  & \texttt{gemini-embedding-2} (3072d)   & --- & 0.938 & 0.900 & 0.919 \\
3  & \textbf{UTE-1 (ours)}                 & \checkmark & \bestopen{0.916} & \bestopen{0.863} & \bestopen{0.889} \\
4  & \texttt{Qwen3-Embedding-4B}           & \checkmark & 0.893 & 0.795 & 0.844 \\
5  & \texttt{arctic-embed-l-v2.0}$^{\ast}$ & \checkmark & 0.876 & 0.850 & 0.863 \\
6  & \texttt{multilingual-e5-large}        & \checkmark & 0.832 & 0.763 & 0.797 \\
7  & \texttt{multilingual-e5-large-inst.}  & \checkmark & 0.815 & 0.811 & 0.813 \\
8  & \texttt{jina-embeddings-v5-small}     & \checkmark & 0.803 & 0.777 & 0.790 \\
9  & \texttt{multilingual-e5-base}         & \checkmark & 0.753 & 0.718 & 0.735 \\
10 & \texttt{Qwen3-Embedding-0.6B}         & \checkmark & 0.685 & 0.743 & 0.714 \\
11 & \texttt{harrier-oss-v1-0.6b}          & \checkmark & 0.635 & 0.758 & 0.696 \\
12 & \texttt{harrier-oss-v1-270m}          & \checkmark & 0.612 & 0.756 & 0.684 \\
13 & \texttt{geevec-embeddings-1.0-lite}   & \checkmark & 0.528 & 0.506 & 0.517 \\
14 & \texttt{embeddinggemma-300m}          & \checkmark & 0.489 & 0.695 & 0.592 \\
\bottomrule
\end{tabular}
\caption{Embedder retrieval quality on our in-domain legal benchmark (top-5
accuracy over 178 queries, \S\ref{sec:retrbench}) and the Uzbek MTEB subset
(nDCG@5 over 127 queries / 1{,}263 candidates), sorted by
the in-domain metric. \best{Bold} = best overall; \bestopen{underline} = best
open-weight. $^{\ast}$Base model for UTE-1: fine-tuning lifts in-domain top-5
from 0.876 to 0.916 ($+$0.040) and MTEB from 0.850 to 0.863, closing roughly half of the remaining gap to the proprietary leader.}
\label{tab:embed}
\end{table*}

\begin{table*}[t]
\centering
\small
\setlength{\tabcolsep}{5pt}
\begin{tabular}{@{}llcccccc@{}}
\toprule
\textbf{Generator} & \textbf{Open} & \textbf{Retriever} & \textbf{Overall} &
\textbf{Compl.} & \textbf{Ctx.\ Acc.} & \textbf{Halluc.-Free} & \textbf{Legal Compr.} \\
\midrule
\texttt{gemini-3-flash} $^{\bullet}$ & --- & \texttt{gemini-001} & \best{0.850} & \best{0.831} & \best{0.889} & 0.772 & \best{0.907} \\
\texttt{claude-sonnet-4-6} $^{\circ}$ & --- & \texttt{gemini-001} & 0.839 & 0.758 & 0.875 & 0.869 & 0.853 \\
\texttt{gemini-3.1-flash-lite} $^{\bullet}$ & --- & \texttt{gemini-001} & 0.817 & 0.653 & 0.885 & 0.911 & 0.821 \\
\texttt{gpt-5.4} $^{\circ}$ & --- & \texttt{gemini-001} & 0.809 & 0.659 & 0.815 & \best{0.990} & 0.772 \\
\texttt{gemma-4-31b-it} $^{\star}$ & \checkmark & \texttt{gemini-001} & 0.803 & 0.679 & 0.849 & 0.899 & 0.786 \\
\texttt{gemma-4-31b-it} $^{\star}$ & \checkmark & \texttt{UTE-1} & 0.724 & 0.534 & 0.788 & 0.879 & 0.694 \\
\texttt{qwen3.6-35b-a3b} $^{\star}$ & \checkmark & \texttt{gemini-001} & 0.721 & 0.661 & 0.774 & 0.687 & 0.762 \\
\textbf{\texttt{qwen3.6-35b-a3b}} $^{\star}$ & \checkmark & \textbf{UTE-1} & 0.671 & 0.563 & 0.728 & 0.710 & 0.681 \\
\texttt{gemma-4-26b-a4b-it} $^{\star}$ & \checkmark & \texttt{gemini-001} & 0.665 & 0.520 & 0.687 & 0.810 & 0.643 \\
\texttt{grok-4.3} $^{\circ}$ & --- & \texttt{gemini-001} & 0.593 & 0.333 & 0.635 & 0.929 & 0.476 \\
\texttt{gemma-4-26b-a4b-it} $^{\star}$ & \checkmark & \texttt{UTE-1} & 0.570 & 0.397 & 0.589 & 0.784 & 0.510 \\
\bottomrule
\end{tabular}
\caption{End-to-end QA quality (\S\ref{sec:qabench}), scored on all 504 items by the third-family judge \texttt{meta/muse-spark-1.1}: no Meta model appears in the candidate set or deployed pipeline. Rows are ordered by Muse overall score. $^{\bullet}$in-budget cloud candidate; $^{\circ}$frontier API shown
as a reference point but excluded by the cost ceiling; $^{\star}$open-weight, on-premises
candidate. The \best{deployed} stacks are
\texttt{gemini-3-flash}$+$\texttt{gemini-001} (cloud) and \texttt{qwen3.6-35b-a3b}$+$UTE-1 (on-premises, see \S\ref{sec:onprem}).}
\label{tab:qa}
\end{table*}

\section{On-Premises Deployment: Sovereignty, Open Weights, and Latency}
\label{sec:onprem}

The on-premises regime removes API models entirely. Selection becomes: among open-weight models that fit the client's hardware, maximize quality subject to a latency budget. We address the retriever and generator in turn, and then explain why we did \emph{not} fine-tune the generator.

\subsection{Closing the retriever gap by fine-tuning (UTE-1)}
\label{sec:ute}
Table~\ref{tab:embed} shows that the best open embedder out of the box
(\texttt{arctic-embed-l-v2.0}) trails the proprietary leader by a meaningful
margin on our in-domain benchmark (0.876 vs.\ 0.961 top-5). Given the asymmetry
argued in Section~\ref{sec:system}, we
judged this gap worth closing, and embedders are cheap to adapt relative to LLMs.
We fine-tuned \texttt{arctic-embed-l-v2.0} into \textbf{UTE-1} from two data sources: (i) $\approx$10k noisy synthetic translations of community QA data, following a recipe of \citet{navasardyan2026less} for adapting embedders to low-resource languages with deliberately imperfect supervision; and (ii) $\approx$7k question--chunk pairs mined from positively-rated production interactions of the cloud service, providing in-domain legal positives. An ablation isolates the two sources: training on the 10k synthetic pairs \emph{alone} already yields a strong open embedder (in-domain top-5 {0.90}), and adding the 7k production-mined real pairs improves it further to 0.916 --- our final UTE-1, which is state-of-the-art among open embedders on both our benchmark and MTEB[UZ] (Table~\ref{tab:embed}), $+$0.040 top-5 over its base
and closing roughly half the in-domain gap to the proprietary leader. This is the empirical core
of our ``invest in the retriever'' thesis: a cheap fine-tune with synthetic data plus a little real in-domain signal recovers roughly half of the retrieval quality that separates open from proprietary.

\subsection{Choosing the generator: quality, then latency}
\label{sec:onpremgen}
Among open generators (Table~\ref{tab:qa}), \texttt{gemma-4-31b-it} is the strongest open model overall and \texttt{qwen3.6-35b-a3b} the second (0.724 vs.\ 0.671 with the UTE-1 retriever). We nonetheless deploy the Qwen model, because the on-premises objective is quality \emph{subject to a latency budget}, and the two models sit at very different points on the compute frontier: \texttt{gemma-4-31b} is a dense model that activates all 31B parameters per token, whereas \texttt{qwen3.6-35b-a3b} is a mixture-of-experts model \citep{fedus2022switch} that activates only $\approx$3B parameters per token.

We profiled the top 2 open candidates on a single-tenant H200 hardware over the 504-query benchmark (Table~\ref{tab:latency}). Both models are inferences using same vLLM (\cite{kwon2023efficientmemorymanagementlarge}) setup paired with their official MTP speculative decoding drafter models. Results show, that the 3B-active Qwen model produces both the first token (TTFT p50) as well as the full response (Total p50) faster than dense Gemma while producing more reasoning tokens.

This matters when serving on-premises as the 3B-active model sustains higher throughput and more concurrent users at lower cost than the 31B-dense one - a
large efficiency gain for a modest quality concession. We therefore deploy \texttt{qwen3.6-35b-a3b} on-premises.

\begin{table}[t]
\centering
\small
\setlength{\tabcolsep}{4pt}
\begin{tabular}{@{}lccccc@{}}
\toprule
\textbf{Model} & \makecell[c]{\textbf{TTFT} \\ \textbf{p50}} & \makecell[c]{\textbf{Total} \\ \textbf{p50}} & \makecell[c]{\textbf{Decode} \\ \textbf{tok/s}} & \makecell[c]{\textbf{Out tok} \\ \textbf{p50}} \\

\midrule
\textbf{\texttt{qwen3.6-35b-a3b}} & \textbf{2.4 s}  & \textbf{29.7 s} & \textbf{171} & 4944 \\
\texttt{gemma-4-31b}              & 14.7 s          & 63.9 s          & 18           & 736  \\
\bottomrule
\end{tabular}
\caption{Latency of the top two open candidates over the 504-query benchmark, measured on a single-tenant H200 hardware. MoE Qwen model is favored by all 3 latency measures while producing more reasoning tokens.}
\label{tab:latency}
\end{table}

\subsection{Why we did \emph{not} fine-tune the generator (a negative result)}
\label{sec:negative}
A natural next step is to fine-tune the open generator on Uzbek legal QA. Under
our resource constraints the only feasible method is QLoRA
\citep{dettmers2023qlora,hu2022lora}. We attempted to QLoRA-fine-tune an open MoE
model (\texttt{gemma-4-26b-a4b}) on 8 AMD GPUs, and abandoned the effort for
reasons we think are instructive.

\emph{It is impractical.} Faithful RAG training requires conditioning on the
retrieved context, but our contexts run to 55--90k tokens (\S\ref{sec:system}).
The obstacle is not a hard capacity wall---one can always trade batch size,
sequence packing, or checkpointing for memory---but economics: every training and
evaluation step processes these very long sequences, turning each experiment into
a slow, expensive run and the hyperparameter search into a low-throughput loop we
could not justify. The tempting shortcut---dropping the context to shorten
inputs---trains the model to answer legal questions \emph{without} retrieved
provisions, exactly the behavior that causes hallucination in deployment, so it is
not an acceptable fix.

\emph{It is also unnecessary.} Even if we could train it, what would it teach?
(i) Better Uzbek: plausibly out of reach for a light QLoRA adapter and, more
importantly, a depreciating investment, since base multilingual LLMs improve
release over release. (ii) The content of the law: this would require continual
pretraining, not a small adapter---and laws are amended, so any weights-baked
legal knowledge is stale on arrival. The correct abstraction for changing law is
retrieval, not memorization. This is the second half of our thesis: the
generator's gap is better addressed by \emph{renting or swapping} a stronger
model than by fine-tuning, whereas the retriever's gap is worth closing once and
reusing. As multilingual base models keep improving release over release, the
case for adapter-tuning a generator for this task only weakens further.

\subsection{Reducing latency further}
\label{sec:specdec}
Since latency is the binding on-premises constraint, it is where we spend
additional effort. Both candidate open families natively support
\emph{multi-token prediction} (MTP), a self-speculative scheme in which auxiliary
heads propose future tokens the model verifies in parallel
\citep{leviathan2023speculative,chen2023speculativesampling}. We serve
\texttt{qwen3.6-35b-a3b} with its MTP head drafting two tokens per step, a further
$\approx$1.2$\times$ decode speedup at no quality change---stacking with the
active-parameter advantage.

\section{Discussion: Lessons Learned}
\label{sec:discussion}

\paragraph{In low-resource RAG, the retriever is the higher-leverage
investment.} The open/proprietary gap was smaller for retrieval than generation
and far cheaper to close---a modest embedder fine-tune (\S\ref{sec:ute}) recovered
roughly half of it, while generator fine-tuning was costly and of dubious value
(\S\ref{sec:negative})---so on a limited budget, spend it on the embedder first.

\paragraph{Build domain evaluation before choosing models, and validate the judge.} General leaderboards did not predict in-domain legal quality, and our entire selection rests on an automated judge. Spending first on a small, expert-grounded benchmark and a human-validated judge protocol (Section~\ref{sec:judge}) was the precondition for every subsequent decision.

\paragraph{Digraphia is a concrete hazard general benchmarks miss.} Uzbek is
written in both Latin and Cyrillic, and our deployment surfaced \emph{script
switching} (e.g. a Cyrillic question drawing a Latin answer, or one answer mixing
scripts) driven by Latin-skewed pretraining and a mixed-script corpus whose
retrieved context can span both scripts (23 benchmark items). 

The impact is more severe in citations. Script switching inside a law title or article heading can break or misdirect a citation. This affects 5.08\% of law citations pre-normalization. Given that citation of the relevant legal articles is a critical component of the product, we post-process results with a deterministic rule-based Latin$\leftrightarrow$Cyrillic transliteration applied to Uzbek spans only (Russian text which also uses Cyrillic script is untouched). The post-processing is content-preserving and resolves all such observed cases fixing broken citations.

\paragraph{Sovereignty is a first-class constraint---and it costs observability.}
For legal and public-sector clients, ``open weights, on-premises'' is a
requirement, not a preference, and it is what made the open-model and latency work
necessary. It also withholds telemetry: the cloud service gives an online signal
(\S\ref{sec:cloud}) that agrees with the offline ranking, but on-premises
deployments return none---the data that cannot leave cannot be measured---raising
the stakes on getting offline evaluation right before shipping.

\section{Conclusion}
We described a production Uzbek legal RAG assistant deployed across cloud and
on-premises regimes, where two benchmarks and a validated judge drove principled
model selection under cost, sovereignty, and latency constraints. The transferable
lesson: in low-resource legal RAG, invest in the retriever (fine-tune once, reuse)
and rent or swap the generator rather than fine-tune it.

\section*{Limitations}
Our study covers a single language (Uzbek), a single domain (statutory law), and a single jurisdiction; the specific model rankings will not transfer, though we believe the decision procedure and the retriever-first lesson are more general. The benchmarks are modest in size (178 retrieval queries; 504 QA items) and the judge-validation set is small (52 items); we therefore avoid over-interpreting small score differences and will release the data so others can extend it. Each item was annotated by a single domain expert, so we cannot report inter-annotator agreement; we mitigate this by using qualified legal professionals and by cross-validating the automated judge against both human scores and a family-independent judge (\S\ref{sec:judge}), but the lack of multiply-annotated items is a genuine limitation. Our headline quality metric is produced by an LLM judge; although we chose an independent judge and validated it against humans, potential bias (e.g., shared pretraining data or stylistic preferences not captured by our held-out set) cannot be ruled out, and the human study itself is limited in scale. The negative QLoRA result is specific to our hardware, our context lengths, and the methods we could afford; it shows that this path was not worth the cost of iteration \emph{for us}, not that generator fine-tuning is infeasible in general. Model identifiers refer to versions current at the time of writing; given the pace of releases, absolute numbers will date quickly, and several reported latency, cost, and judge-correlation figures are deployment-specific and should be read as characterizing our setting rather than as universal constants.

\section*{Acknowledgements}
The authors would like to express their gratitude to the BalcomSoft and Tuzuk.AI teams, with special thanks to Avazbek Nuriddinov and Askar Djumanov for their invaluable support. This work was made possible through their generous provision of data, domain expertise, and high-quality annotations.

\section*{Ethical Considerations}
The system answers legal questions but is an information-retrieval aid, not a
substitute for qualified legal counsel; deployed interfaces present answers with
citations to source provisions and with disclaimers to that effect, and we
caution against using outputs as definitive legal advice. Hallucination is an
acute risk in this domain, which is precisely why we (i) ground answers in
retrieved statute, (ii) measure a dedicated hallucination-free dimension, and
(iii) chose RAG over weights-baked legal knowledge so that answers track the
current law. The on-premises offering exists to honor data-sovereignty and
confidentiality requirements of legal and public-sector clients, keeping
sensitive documents on client infrastructure. Production interaction data used to
build training sets (Section~\ref{sec:ute}) was limited to positively-rated
exchanges and handled under the product's terms of service; we use it in
aggregate to mine question--provision relevance, not to expose individual users.
Expert annotators contributing reference answers and human judgments were
compensated domain professionals. Because the legal corpus and the assistant
operate in a specific national context, deployment elsewhere would require
re-grounding in the relevant law.

\bibliography{custom}

\appendix

\section{UzLib Leaderboard (Candidate Pre-Filtering)}
\label{app:uzlib}
We use the public UzLib leaderboard of general Uzbek language ability as the
first of three pre-filters on LLM candidates (Section~\ref{sec:cloud}). UzLib
measures generic language ability rather than grounded legal QA, so we treat it
only as a coarse admissibility filter, not as a selection metric; final selection
uses our end-to-end QA benchmark (Table~\ref{tab:qa}). The full leaderboard,
abbreviated to the entries relevant to our candidate set, is reproduced from the
public source; we omit it here for space and will include it in the camera-ready
appendix.

\section{Cost--Quality Analysis}
\label{app:cost}
This appendix expands the cost--quality result summarized in
Section~\ref{sec:cloud}. For each generator we measured the cost of answering the
full 504-query QA benchmark through each provider's standard (non-batch) API
(Table~\ref{tab:cost}); standard pricing is what production incurs. We use
\emph{inference} cost as the deployment-relevant axis. The cost of the LLM judge
is a separate, roughly constant evaluation overhead ($\approx$\$22 per model,
since the same judge scores every model's 504 answers) and is excluded from the
deployment comparison.

\begin{figure}[t]
\centering
\includegraphics[width=0.85\linewidth]{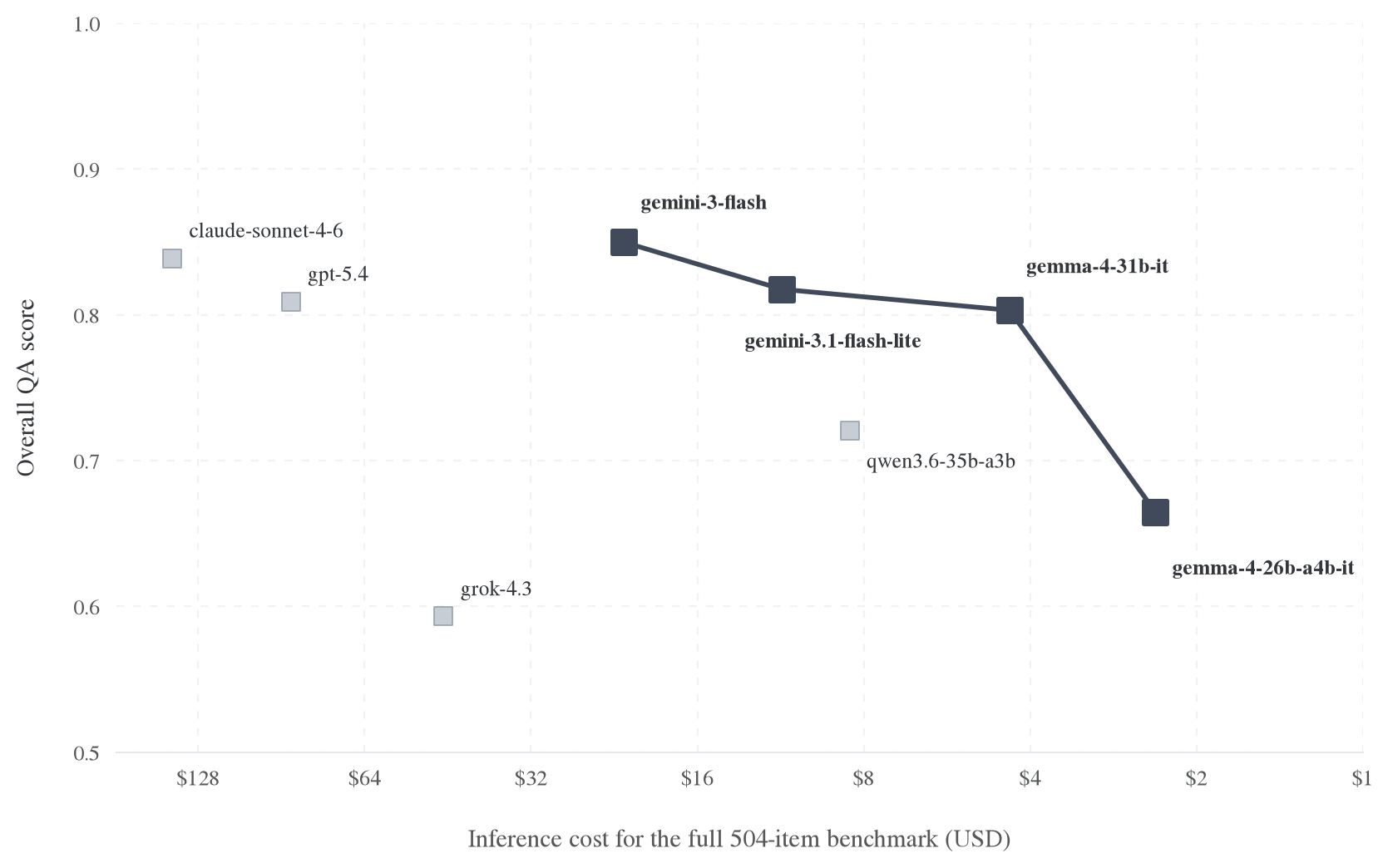}
\caption{Cost--quality landscape (Appendix~\ref{app:cost}). Top-right is better. \texttt{gemini-3-flash} achieves highest quality on the Pareto frontier. Among open models, the on-premises pick
(\texttt{qwen3.6-35b-a3b}) is chosen on serving throughput, not this hosted-API
cost axis (\S\ref{sec:onpremgen}).}
\label{fig:pareto}
\end{figure}

Figure~\ref{fig:pareto} plots inference cost (log scale) against overall QA score
(\texttt{gemini-001} retriever, for a common comparison). The Pareto frontier
runs through \texttt{gemma-4-26b-a4b}, \texttt{gemma-4-31b},
\texttt{gemini-3.1-flash-lite}, and \texttt{gemini-3-flash}. Two readings matter.

\emph{For the cloud regime}, \texttt{gemini-3-flash} is the quality-maximizing
vertex of the frontier and \emph{strictly dominates} all three frontier APIs:
each costs more \emph{and} scores lower (\texttt{grok-4.3} $2.1\times$,
\texttt{gpt-5.4} $4.0\times$, \texttt{claude-sonnet-4-6} $6.6\times$ the cost).
This is the empirical backbone of ``rent the flash tier, not the frontier.''

\emph{For the on-premises regime}, this cost axis does \emph{not} carry over. Per-token
hosted pricing reflects cloud economics; on owned hardware the binding cost is
serving throughput, governed by active parameters
(\S\ref{sec:onpremgen})---which is why the deployed on-premises model
(\texttt{qwen3.6-35b-a3b}, 3B active) differs from the hosted-API cost--quality
pick among open models (\texttt{gemma-4-31b}). The figure is a cloud-regime tool;
the same ``hosted metrics mislead on-prem'' caveat from latency
(\S\ref{sec:onpremgen}) applies to cost.

\begin{table}[t]
  \centering
  \small
  \setlength{\tabcolsep}{4pt}
  \begin{tabular}{@{}lrr@{}}
    \toprule
    \textbf{Generator} & \textbf{QA} & \textbf{Inf. cost (\$)} \\
    \midrule
    \texttt{gemini-3-flash}$^{\bullet}$         & \textbf{0.850} &  21.72 \\
    \texttt{claude-sonnet-4-6}$^{\circ}$       & 0.839          & 142.63 \\
    \texttt{gemini-3.1-flash-lite}$^{\bullet}$ & 0.817          &  11.24 \\
    \texttt{gpt-5.4}$^{\circ}$                 & 0.809          &  87.02 \\
    \texttt{gemma-4-31b-it}$^{\star}$          & 0.803          &   4.36 \\
    \texttt{qwen3.6-35b-a3b}$^{\star}$         & 0.721          &   8.48 \\
    \texttt{gemma-4-26b-a4b-it}$^{\star}$      & 0.665          &   2.37 \\
    \texttt{grok-4.3}$^{\circ}$                & 0.593          &  46.05 \\
    \bottomrule
  \end{tabular}
  \caption{Overall QA score with the \texttt{gemini-001} retriever, scored on all 504 queries by the third-family judge \texttt{meta/muse-spark-1.1}, and generator-inference cost at each provider's standard list rate.}
  \label{tab:cost}
\end{table}

\section{Reproducibility Details}
\label{app:repro}
\paragraph{Production pipeline.} Chunks are segmented at provision granularity and
prepended with their hierarchical headings (code/chapter/article) to preserve
context; oversized statutory tables are indexed as multiple chunks but re-expanded
in full whenever any chunk is retrieved. Retrieval is hybrid (dense $+$ lexical),
preceded by question normalization and, where needed, translation; answers are
post-processed to normalize script (Latin/Cyrillic) to match the question
(\S\ref{sec:discussion}). Training and evaluation sets are de-duplicated with both
semantic and fuzzy lexical matching.

\end{document}